\documentclass[letterpaper]{article} 
\usepackage{aaai2026}  
\usepackage{times}  
\usepackage{helvet}  
\usepackage{courier}  
\usepackage[hyphens]{url}  
\usepackage{graphicx} 
\usepackage{subcaption}
\usepackage{natbib}  
\usepackage{caption} 
\usepackage{algorithm}
\usepackage{algorithmic}

\usepackage{newfloat}
\usepackage{listings}
\DeclareCaptionStyle{ruled}{labelfont=normalfont,labelsep=colon,strut=off} 
\floatstyle{ruled}
\newfloat{listing}{tb}{lst}{}
\floatname{listing}{Listing}

\usepackage{url}
\usepackage{booktabs}
\usepackage{amsmath}
\usepackage{amssymb}
\usepackage{multirow}
\usepackage{makecell}
\usepackage{colortbl}
\usepackage{xcolor}
\usepackage{tcolorbox}
\usepackage{xspace}

\usepackage{threeparttable}
\usepackage{pifont}

\title{How Good LLMs Are at Answering Bangla Medical Visual Questions?\\Dataset and Benchmarking}
\author {
    Rafid Ahmed\textsuperscript{\rm 1,*},
    Intesar Tahmid\textsuperscript{\rm 1,*},
    Mir Sazzat Hossain\textsuperscript{\rm 2,*}, \\
    Tasnimul Hossain Tomal\textsuperscript{\rm 1},
    Md Fahim\textsuperscript{\rm 1, 2, \dag},
    Md Farhad Alam Bhuiyan\textsuperscript{\rm 1}
}
\affiliations {
    \textsuperscript{\rm 1}Penta Global Limited, Bangladesh\\
    \textsuperscript{\rm 2}Center for Computational \& Data Sciences, Independent University, Bangladesh\\
    \textsuperscript{*}\small{Equal Contribution}, \quad \textsuperscript{\dag}\small{Project Lead}\\
    \small{\textbf{Correspondence:} \texttt{\{ahmedrafid023, intesar3006, pdcsedu\}}{@gmail.com}
 }
}

\begin{document}

\maketitle

\begin{abstract}
Recent advancements in Large Language Models (LLMs) and Large Vision Language Models (LVLMs) have enabled general-purpose systems to demonstrate promising capabilities in complex reasoning tasks, including those in the medical domain. Medical Visual Question Answering (MedVQA) has particularly benefited from these developments. However, despite Bangla being one of the most widely spoken languages globally, there exists no established MedVQA benchmark for it. To address this gap, we introduce BanglaMedVQA, a dataset comprising clinically validated image–question–answer pairs, along with a comprehensive evaluation of current foundation models on this resource. Consistent with prior findings that report low performance of current models on English MedVQA benchmarks, our analysis reveals that Bangla performance is substantially lower, reflecting the challenges inherent to low-resource languages. Even top-performing models such as Gemini and GPT-4.1 mini fail to accurately answer specialized diagnostic questions, indicating severe limitations in fine-grained medical reasoning. Although certain open-source models, such as Gemma-3, occasionally outperform these models in general categories, they too struggle with clinically complex questions, underscoring the urgent need for top-notch evaluation method.
\end{abstract}

\section{Introduction}
In recent years, foundation models such as Large Language Models (LLMs) \cite{achiam2023gpt, touvron2023llama, anil2023palm} and LVLMs\cite{gemini2025, li2023blip, Liu2023ImprovedBW, Chen2023MiniGPTv2LL} have attracted significant attention for their ability to process, generate, and reason over complex textual and visual inputs. These models produce human-like language and achieve high performance across a wide range of benchmarks. Their integration into the medical domain has already shown promising results, demonstrating potential for real-world clinical applications.

\begin{figure*}[h]  
    \centering
    \includegraphics[width=0.98\linewidth]{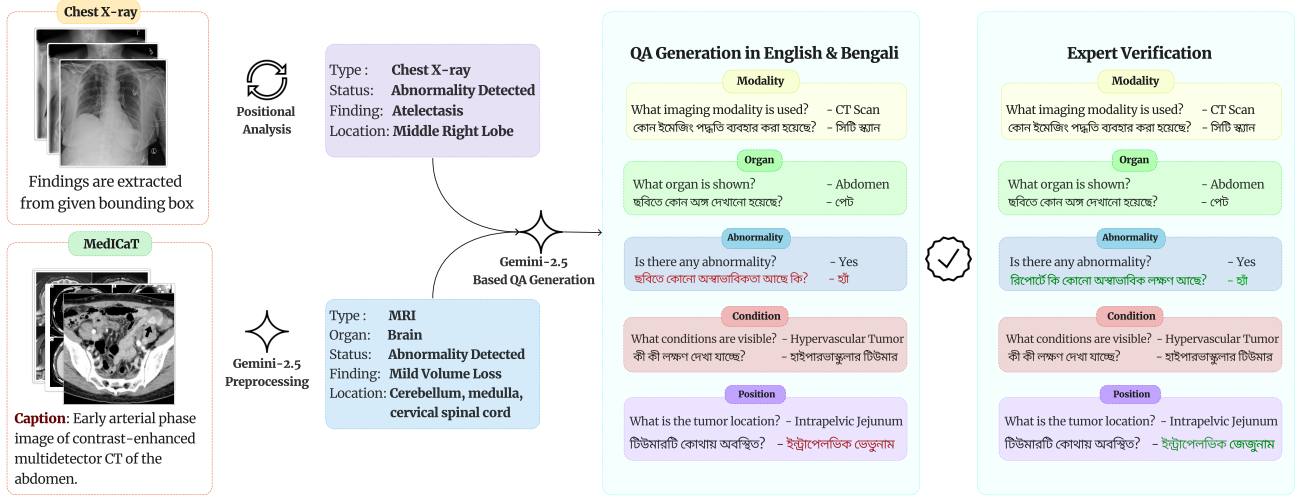}  
    \caption{Workflow of the dataset curation process. Images and metadata were obtained from two widely used biomedical datasets, enabling the automatic generation of QA pairs, which were subsequently verified by domain experts.}

    \label{fig:workflow}
\end{figure*}

Applications such as automated radiology report generation \cite{sloan2024automated}, clinical decision support\cite{poulain2024bias}, and interactive medical question answering\cite{kim2024medexqa} highlight the capability of LLMs to assist both patients and healthcare professionals. Existing medical VQA benchmarks \cite{liu2021slake, he2020pathvqa, zhang2023pmc}, when combined with current foundation models, have reported encouraging results. However, despite being the seventh most spoken language in the world, with approximately 278 million speakers, Bangla has received very limited attention in the context of medical visual question answering. It is therefore crucial to evaluate the performance of LLMs in this domain for Bangla.

The effectiveness of medical AI systems is closely linked to the quality of the datasets on which they are trained \cite{gong2023survey}. In high-resource languages such as English, medical VQA datasets have gradually evolved in complexity, progressing from simple factual questions to tasks requiring deeper clinical reasoning \cite{marino, schwenk2022okvqa}. In contrast, low-resource languages face a significant bottleneck: the absence of sufficiently comprehensive datasets capable of supporting even fundamental diagnostic question answering.

In the Bangla medical VQA domain, progress has been highly limited. To date, the only publicly available dataset is \cite{medvqabn}. However, this resource does not provide any information regarding the source of the images, nor does it describe the process of dataset creation or validation by medical experts. Furthermore, it suffers from annotation errors and lacks clinical oversight, making it unsuitable as a reliable benchmark.

To address this gap, we propose a new Bangla medical Visual Question Answering (VQA) dataset built from images drawn from multiple medical domains \cite{Subramanian2020MedICaTAD, Wang2017ChestXRay8HC}, paired with diverse and clinically relevant questions. The dataset contains 1,374 unique image–caption pairs, with questions spanning five categories: modality, organ, abnormality, condition, and position. These categories can be grouped into two types: (i) generalized questions, which are relatively straightforward, and (ii) specialized questions, which require deeper medical understanding. Importantly, unlike the existing resource, our dataset was validated by two certified physicians to ensure both clinical accuracy and reliability.

We systematically evaluate both open-source and closed-source LLMs on our proposed dataset. The best-performing models, Gemini and GPT, achieve overall accuracies of 40.38\% and 26.50\%, respectively. While results on generalized questions appear promising, performance on specialized diagnostic categories such as \textit{Condition/Finding} and \textit{Position} falls below random chance, revealing critical gaps in the reliability of current LVLMs for fine-grained medical reasoning. Previous studies have reported worse-than-random behavior on English medical VQA tasks \cite{yan2024worse}; however, the problem is even more pronounced for Bangla, a low-resource language. Comparative experiments further confirm that closed-source models consistently perform better in English than in Bangla.

Incorporating chain-of-thought reasoning and supplementing inputs with visual descriptions generated by GPT-4.1 mini substantially improves performance, suggesting that limited visual understanding remains a key bottleneck. These results highlight the importance of augmenting LVLMs with richer visual and linguistic context to enable more reliable medical VQA in low-resource languages. These findings highlight the urgent need for more robust Bangla medical datasets and stronger evaluation frameworks, despite the impressive progress of foundation models in general-domain benchmarks. In summary, our contributions are:
\begin{itemize}
    \item We introduce a new Bangla Medical VQA dataset containing diverse question–answer pairs across diagnostic categories for medical images.
    \item We provide a comprehensive benchmark that systematically evaluates open and closed-source VLMs under zero-shot and Chain-of-Thought conditions.
\end{itemize}

 \section{Related Work} \label{sec:related_work}

\subsection{Visual Question Answering}
Visual Question Answering (VQA) has emerged as a challenging task at the intersection of vision and language. It was formally established by \citet{Antol_2015_ICCV}, who introduced a benchmark dataset requiring fine-grained visual understanding and commonsense reasoning. \citet{Shih_et_al_CVPR} further advanced the field by introducing a region-focused attention mechanism, aligning image regions with question embeddings to improve performance on queries requiring localized reasoning. However, such early models exploited dataset biases, which led \citet{Goyal_2017_CVPR} to propose VQA v2.0, a balanced dataset with complementary image-question pairs that force models to genuinely leverage visual content. Beyond visual grounding, \citet{Sanket_et_al_2019} addressed knowledge-intensive scenarios with KVQA, a dataset requiring external knowledge, particularly about named entities, to answer complex questions. Together, these works have progressively shaped VQA from simple recognition toward deeper reasoning.

\begin{figure*}[t]
    \centering
    \hfill
    \begin{subfigure}[b]{0.4\textwidth}
    \includegraphics[width=\textwidth]{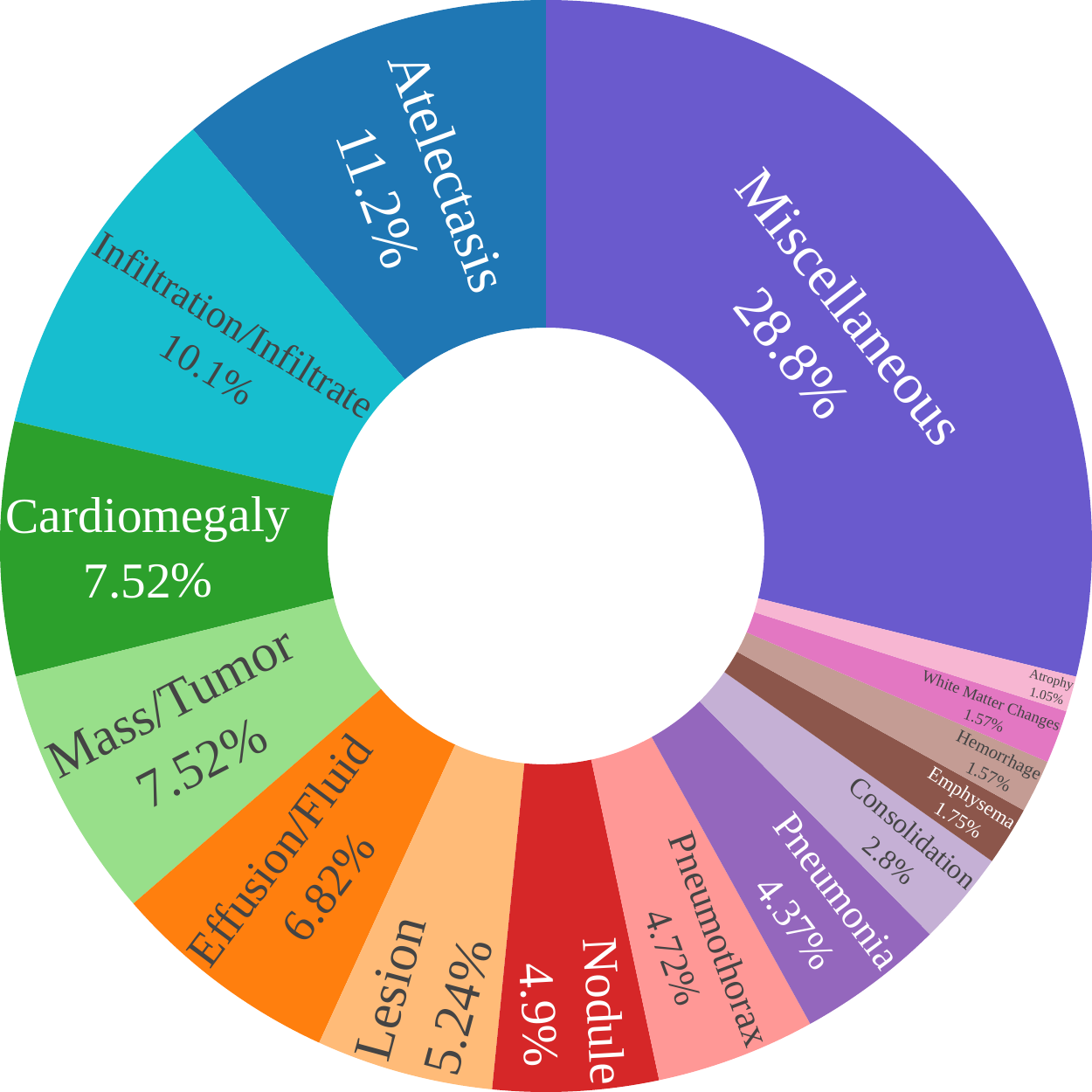}
    \caption{Distribution of Clinical Conditions in the Dataset.}
    \label{fig:condition_distribution}
    \end{subfigure}
    \hfill
    \begin{subfigure}[b]{0.38\textwidth}        
    \includegraphics[width=\textwidth]{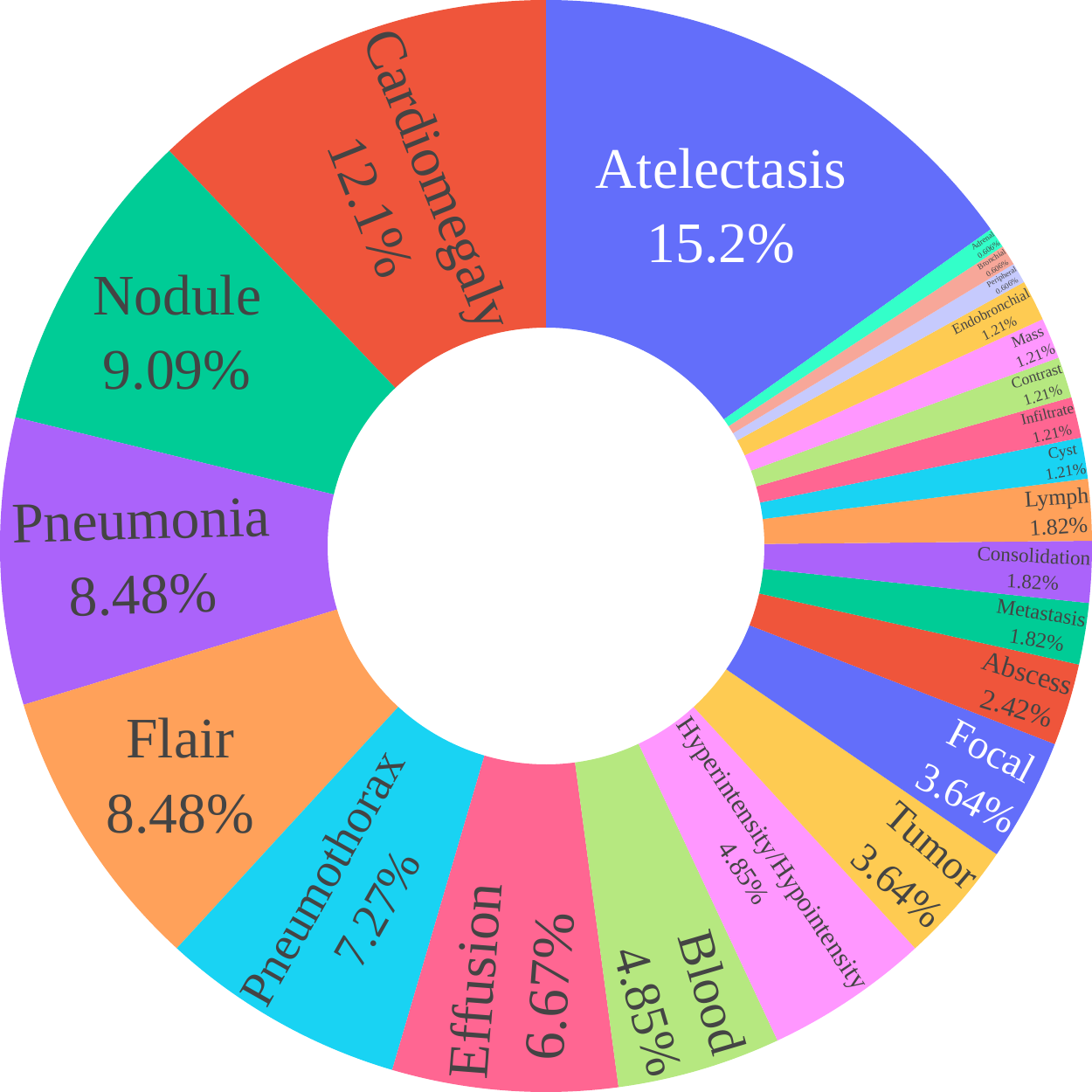}
    \caption{Distribution of Question Keywords in the Dataset.}
    \label{fig:key_distribution}
    \end{subfigure}
    \label{fig:dist_donut}
    \caption{Distributions of clinical conditions and question keywords in the curated dataset.}
\end{figure*}

\subsection{Medical Visual Question Answering}
\citet{Gu_2024_CVPR} proposes a Latent Prompt Assist architecture for medical VQA that generates answer-constrained latent prompts and fuses them with unimodal and multimodal features. It also integrates disease–organ priors to enhance clinical relevance, and ultimately, it reported state-of-the-art gains on VQA-RAD \cite{VQA-RAD}, SLAKE\cite{liu2021slake}, and VQA-2019 \citet{Yim_DermaVQA_MICCAI2024} delivers a multilingual dermatology VQA dataset and benchmarks for consumer-generated images and free-text clinician responses, emphasizing real-world noise, longer responses, and multilingual evaluation metrics. \citet{xu-etal-2024-mlevlm} proposed a multi-level visual language model for MVQA, introducing a new multi-level instruction dataset (MLe-VQA), a feature alignment module, and an evaluation benchmark (MLe-Bench) to better capture recognition, details, diagnosis, knowledge, and reasoning. \citet{gai-etal-2025-medthink} developed rationale-guided benchmarks (R-RAD, R-SLAKE, R-Path) and a framework that integrates medical decision-making rationales into MedVQA, thereby improving both interpretability and performance. \citet{Yu_Tong_Yu_Zhang_2025} introduced adaptive region-level visual prompts and a hierarchical answer generator with PEFT techniques to improve fine-grained localization and generative MedVQA.

\subsection{Bangla Visual Questing Answering}
\citet{DBLP:journals/corr/abs-2410-14991} introduced a regionally relevant Bangla VQA corpus named ChitroJera that emphasizes cultural and contextual relevance for Bangla speakers, filling important gaps left by predominantly English VQA resources. Bangla-Bayanno by \citet{hasan2025banglabayanno52kpairbengalivisual} presents a large-scale, open-ended Bangla VQA benchmark (52.7K QA pairs over \~4.7K images) created via an LLM-assisted translation-refinement pipeline to produce fluent, context-preserving Bangla queries and answers. \citet{10055205} introduced the first human-annotated Bengali VQA dataset using images from VQA v2.0 and proposed a deep learning-based top-down attention model to effectively balance visual and linguistic information in QA tasks.  

\section{Dataset Creation} \label{sec:dataset}

\subsection{Data Collection}
Our primary objective was to evaluate the performance of LVLMs on medical Visual Question Answering (VQA) in Bangla. To ensure we were testing models against clinically relevant and large-scale data, we decided not to build the entire dataset from scratch. Instead, we leveraged the proven scale and complementary strengths of two widely used, validated biomedical datasets: MedICaT~\cite{Subramanian2020MedICaTAD} and ChestX-ray14~\cite{Wang2017ChestXRay8HC} to construct the BanglaMedVQA. We selected these datasets because they provide complementary forms of clinically rich information essential for assembling high-quality VQA pairs. Specifically, the MedICaT dataset contains detailed condition information and positional descriptions directly embedded in image--caption pairs, enabling extraction of clinically contextualized findings from textual content. On the other hand, the ChestX-ray14 dataset offers structured condition metadata and explicit positional annotations in the form of bounding boxes, which can be programmatically converted into spatial reasoning descriptions.

Previous research on medical VQA generation and benchmarking~\cite{zhang2023pmc,li2023self,yan2024worse} has demonstrated that few-shot generation and structured QA pipelines can yield clinically valid and semantically aligned question--answer pairs. Following this line of evidence, we adopted a similar methodological approach to maintain both the clinical integrity and linguistic coherence of our generated data. For each image, a standardized metadata record was created, including the imaging modality, anatomical region, and a list of detected clinical findings along with their positional descriptions. In the case of MedICaT, Gemini was used with few-shot prompting to extract information about abnormalities and their locations from captions. For ChestX-ray14, a specialized positional reasoning module generated descriptive text from bounding box coordinates, as illustrated in Figure~\ref{fig:workflow}.

\subsection{Question Generation from Metadata}
From the curated metadata, we generated English QA pairs for each image using the \texttt{gemini-2.5-flash} model with a few-shot prompting strategy. This method enabled the model to produce clinically coherent and contextually aligned QA pairs consistent with the underlying metadata. Subsequently, to support multilingual accessibility and facilitate research in Bangla clinical contexts, we employed the \texttt{gemini-2.0-flash} model to translate the English QA pairs into Bangla. This two-stage pipeline ensured both linguistic diversity and consistency between the metadata and the QA pairs.

\begin{table*}[t]
    \centering
    \small
    \resizebox{.9\textwidth}{!}{
    \begin{tabular}{lccccccccc}
        \toprule
        \multirow{2}{*}{\textbf{Models}} & \multicolumn{3}{c}{\textbf{Chest X-Ray}} 
        & \multicolumn{3}{c}{\textbf{Medi-CAT}}
        & \multicolumn{3}{c}{\textbf{Overall (Avg.)}}
        \\
        \cmidrule(lr){2-4}
        \cmidrule(lr){5-7}
        \cmidrule(lr){8-10}
      & \textbf{Acc} & \textbf{BScore} & \textbf{LAVE} & \textbf{Acc} & \textbf{BScore} & \textbf{LAVE} & \textbf{Acc} \cellcolor{gray!30} & \textbf{BScore} \cellcolor{gray!30}& \textbf{LAVE} \cellcolor{gray!30}\\    
    \midrule
    \multicolumn{10}{c}{\small \textit{\textbf{Zero Shot Prompt on Bangla QA Pair}}}\\
    \midrule    
    \multicolumn{6}{l}{\small \textit{\textbf{Closed Source VLMs}}}\\
    
    \quad Gemini 2.5 Flash & \textbf{37.00} & \textbf{82.75} & \textbf{50.80} & 34.50 & \textbf{81.14} & \textbf{60.08} & \textbf{35.75} \cellcolor{gray!30}& \textbf{81.95} \cellcolor{gray!30}& \textbf{55.44}\cellcolor{gray!30} \\
    \quad GPT-4.1 mini & 15.00 & 72.93 & 37.70 & 16.00 & 73.51 & 39.50 & 15.50 \cellcolor{gray!30} & 73.22 \cellcolor{gray!30}& 38.60 \cellcolor{gray!30}\\
    
    \multicolumn{6}{l}{\small \textit{\textbf{Open Source VLMs}}}\\
    
    \quad Llama-3.2V 11B & 8.50 & 73.43 & 17.50 & 11.00 & 81.01 & 33.50 & 9.75 \cellcolor{gray!30}& 77.22 \cellcolor{gray!30}& 25.50\cellcolor{gray!30} \\
    \quad Gemma-3 12B & 29.50 & 72.62 & 39.70 & \textbf{39.50} & 76.70 & 53.37 & 34.50 \cellcolor{gray!30}& 74.66 \cellcolor{gray!30}& 46.54 \cellcolor{gray!30}\\
    \quad Qwen2.5-VL 7B & 21.00 & 75.36 & 32.85 & 28.50 & 75.53 & 38.30 & 24.75 \cellcolor{gray!30}& 75.45\cellcolor{gray!30} & 35.58\cellcolor{gray!30} \\
    \quad LLaVA-1.5 7B & 8.50 & 74.43 & 15.20 & 14.50 & 75.14 & 18.75 & 11.50\cellcolor{gray!30} & 74.79\cellcolor{gray!30} & 16.98\cellcolor{gray!30} \\
    
    \multicolumn{6}{l}{\small \textit{\textbf{Open Source Medical VLMs}}}\\
    
    \quad Med-LLaVa 7B & 7.00 & 41.87 & 12.70 & 6.00 & 26.21 & 30.72 & 6.50 \cellcolor{gray!30}& 34.04\cellcolor{gray!30} & 21.71 \cellcolor{gray!30}\\
    \quad Med-Gemma 4B & 4.00 & 68.91 & 11.90 & 18.50 & 76.02 & 27.80 & 11.25 \cellcolor{gray!30}& 72.47 \cellcolor{gray!30}& 19.85 \cellcolor{gray!30}\\
    
    \midrule
    \multicolumn{10}{c}{\small \textit{\textbf{CoT Prompt on Bangla QA Pair}}}\\
    \midrule
    \multicolumn{6}{l}{\small \textit{\textbf{Closed Source VLMs}}}\\
    
    \quad Gemini 2.5 Flash & \textbf{31.00} & \textbf{83.10} & \textbf{49.33} & 34.00 & 82.13 & \textbf{60.58} & 32.50 \cellcolor{gray!30}& \textbf{82.62} \cellcolor{gray!30}& \textbf{54.96} \cellcolor{gray!30}\\
    \quad GPT-4.1 mini & 11.50 & 69.70 & 36.15 & 16.00 & 72.91 & 35.70 & 13.75 \cellcolor{gray!30}& 71.31 \cellcolor{gray!30}& 35.93 \cellcolor{gray!30}\\
    
    \multicolumn{6}{l}{\small \textit{\textbf{Open Source VLMs}}}\\
    
    \quad Llama-3.2V 11B & 13.00 & 73.68 & 30.17 & 21.50 & \textbf{82.94} & 47.18 & 17.25 \cellcolor{gray!30}& 78.31 \cellcolor{gray!30}& 38.68 \cellcolor{gray!30}\\
    \quad Gemma-3 12B & 31.00 & 74.26 & 41.00 & \textbf{40.50} & 78.72 & 56.33 & \textbf{35.75} \cellcolor{gray!30}& 76.49 \cellcolor{gray!30}& 48.67 \cellcolor{gray!30}\\
    \quad Qwen2.5-VL 7B & 20.00 & 74.52 & 31.85 & 29.00 & 75.51 & 37.15 & 24.50\cellcolor{gray!30} & 75.02\cellcolor{gray!30} & 34.50\cellcolor{gray!30} \\
    \quad LLaVA-1.5-7B & 1.30 & 70.44 & 6.05 & 2.00 & 70.58 & 4.55 & 1.65 \cellcolor{gray!30}& 70.51\cellcolor{gray!30} & 5.30\cellcolor{gray!30} \\
    
    \multicolumn{6}{l}{\small \textit{\textbf{Open Source Medical VLMs}}}\\
    
    \quad Med-LLaVa 7B & 2.60 & 15.93 & 14.35 & 3.27 & 31.79 & 32.31 & 2.94 \cellcolor{gray!30}& 23.86 \cellcolor{gray!30}& 23.33 \cellcolor{gray!30}\\
    \quad Med-Gemma 4B & 12.00 & 69.21 & 22.50 & 28.00 & 75.18 & 34.63 & 20.00\cellcolor{gray!30} & 72.20 \cellcolor{gray!30}& 28.57 \cellcolor{gray!30}\\
    
    \bottomrule         
    \end{tabular}
    }
    \caption{Model Benchmarking for \textbf{Bangla} with average scores across Chest X-Ray and Medi-CAT datasets.}
    \label{tab:model_benchmark}
\end{table*}

\subsection{Question and Answer Verification}
The translated QA pairs may exhibit linguistic inconsistencies or positional mismatches between textual descriptions and visual regions. To rigorously evaluate the reliability of the generated Bangla QA pairs, we carried out an expert validation study on the 500-sample subset. Two medical specialists were compensated on an hourly basis to independently examine and verify the annotations. This focused review covered two high-risk areas.

\textbf{1. Multilingual Quality Check:} The Bangla QA subset was subjected to a specialized clinical quality audit to verify the accuracy of technical term translation and check for structural decoding errors, such as missing delimiters, which are common challenges in low-resource clinical translation. 

\textbf{2. Positional Consistency Review:} The specialist review was mandated to allocate focused scrutiny to validating the LLM-inferred positional metadata. This included a visual confirmation process to ensure the textual location details precisely aligned with the pathological region observed in the image, thereby mitigating risks associated with sparse ChestX-ray14 bounding box data.

\begin{table}[t]
\centering
\small
\begin{tabular}{lcc}
\toprule
\multicolumn{3}{l}{\textbf{QA Pair Validation}} \\
\midrule
Samples used & \multicolumn{2}{c}{500} \\
\textbf{Annotator} & \textbf{Acc. (\#)} & \textbf{Acc. (\%)} \\
\midrule
Annotator 1 & 486 & 97.20 \\
Annotator 2 & 488 & 97.60 \\
\midrule
\multicolumn{3}{l}{\textbf{Inter-Rater Reliability (Cohen’s Kappa)}} \\
\midrule
Samples used & \multicolumn{2}{c}{500} \\
Cohen’s $\kappa$ coefficient & \multicolumn{2}{c}{0.89} \\
\bottomrule
\end{tabular}
\caption{Verification statistics of the curated dataset. Accuracy refers to the correctness of QA pairs validated by medical specialists, and Cohen’s $\kappa$ measures inter-rater reliability.}
\label{tab:pilot_verification}
\end{table}

The average acceptance rate across both validators and both quality checks demonstrated an overall accuracy of \textbf{$97\%$}. Based on this high level of verification and the strong inter-rater agreement, the medical specialists provided a favorable verdict, confirming the robustness and clinical reliability of the proposed LLM-driven generation method. Following this successful validation, we proceeded to generate the remaining portion of the dataset using the identical two-stage pipeline, culminating in the final set of 2,000 unique Bangla medical VQA pairs.

\begin{table*}[t]
    \centering
    \small
     \resizebox{.9\textwidth}{!}{
    \begin{tabular}{lccccccccc}
        \toprule
        \multirow{2}{*}{\textbf{Models}} & \multicolumn{3}{c}{\textbf{Chest X-Ray}} 
        & \multicolumn{3}{c}{\textbf{Medi-CAT}}
        & \multicolumn{3}{c}{\textbf{Overall (Avg.)}}
        \\
        \cmidrule(lr){2-4}
        \cmidrule(lr){5-7}
        \cmidrule(lr){8-10}
      & \textbf{Acc} & \textbf{BScore} & \textbf{LAVE} & \textbf{Acc} & \textbf{BScore} & \textbf{LAVE} & \textbf{Acc} \cellcolor{gray!30}& \textbf{BScore} \cellcolor{gray!30}& \textbf{LAVE} \cellcolor{gray!30}\\    
    \midrule
    \multicolumn{10}{c}{\small \textit{\textbf{Zero Shot Prompt on English QA Pair}}}\\
    \midrule    
    \multicolumn{6}{l}{\small \textit{\textbf{Closed Source VLMs}}}\\
        
        \quad Gemini 2.5 Flash & 40.50 & 67.04 & 52.47 & 49.50 & \textbf{73.55} & \textbf{63.15} & 45.00 \cellcolor{gray!30}& \textbf{70.30} \cellcolor{gray!30}& \textbf{57.81} \cellcolor{gray!30}\\
        \quad GPT-4.1 mini & 41.00 & 67.23 & 54.70 & 33.00 & 63.77 & 51.31 & 37.00 \cellcolor{gray!30}& 65.50 \cellcolor{gray!30}& 53.01 \cellcolor{gray!30}\\
        
        \multicolumn{6}{l}{\small \textit{\textbf{Open Source VLMs}}}\\
        
        \quad Llama-3.2V 11B & 10.50 & 59.92 & 45.30 & 11.00 & 66.00 & 53.25 & 10.75 \cellcolor{gray!30}& 62.96 \cellcolor{gray!30}& 49.28 \cellcolor{gray!30}\\
        \quad Gemma-3 12B & 19.50 & 56.05 & 43.30 & 46.50 & 67.14 & 57.60 & 33.00 \cellcolor{gray!30}& 61.60 \cellcolor{gray!30}& 50.45 \cellcolor{gray!30}\\
        \quad Qwen2.5-VL 7B & \textbf{46.50} & \textbf{69.53} & \textbf{55.85} & \textbf{50.00} & 69.44 & 57.10 & \textbf{48.25} \cellcolor{gray!30}& 69.49 \cellcolor{gray!30}& 56.48 \cellcolor{gray!30}\\
        \quad LLaVA-1.5 7B & 17.00 & 45.94 & 23.90 & 40.50 & 59.14 & 47.00 & 28.75 \cellcolor{gray!30}& 52.54\cellcolor{gray!30} & 35.45 \cellcolor{gray!30}\\
    
        \multicolumn{6}{l}{\small \textit{\textbf{Open Source Medical VLMs}}}\\
        
        \quad Med-LLaVa 7B & 1.80 & 37.01 & 32.75 & 2.17 & 31.21 & 39.75 & 1.99 \cellcolor{gray!30}& 34.11 \cellcolor{gray!30}& 36.25\cellcolor{gray!30} \\
        \quad Med-Gemma 4B & 13.00 & 40.32 & 16.70 & 36.00 & 53.84 & 40.00 & 24.50 \cellcolor{gray!30}& 47.08 \cellcolor{gray!30}& 28.35 \cellcolor{gray!30}\\  
    
    \midrule
    \multicolumn{10}{c}{\small \textit{\textbf{CoT Prompt on English QA Pair}}}\\
    \midrule
    \multicolumn{6}{l}{\small \textit{\textbf{Closed Source VLMs}}}\\
        
        \quad Gemini 2.5 Flash & 40.50 & 67.17 & 53.75 & \textbf{48.50} & \textbf{72.32} & \textbf{61.55} & 44.50 \cellcolor{gray!30}& \textbf{69.75} \cellcolor{gray!30}& \textbf{57.65} \cellcolor{gray!30}\\
        \quad GPT-4.1 mini & 33.50 & 62.83 & 46.85 & 46.00 & 68.88 & 58.31 & 39.75 \cellcolor{gray!30}& 65.86 \cellcolor{gray!30}& 52.58 \cellcolor{gray!30}\\
        
        \multicolumn{6}{l}{\small \textit{\textbf{Open Source VLMs}}}\\
        
        \quad Llama-3.2V 11B & 29.00 & 59.21 & 50.12 & 45.50 & 67.68 & 50.25 & 37.25 \cellcolor{gray!30}& 63.45 \cellcolor{gray!30}& 50.19 \cellcolor{gray!30}\\
        \quad Gemma-3 12B & 38.00 & 66.43 & 46.55 & 45.00 & 67.31 & 55.50 & 41.50 \cellcolor{gray!30}& 66.87 \cellcolor{gray!30}& 51.03 \cellcolor{gray!30}\\
        \quad Qwen2.5-VL 7B & \textbf{48.00} & \textbf{69.52} & \textbf{56.55} & 46.00 & 67.82 & 54.50 & \textbf{47.00} \cellcolor{gray!30}& 68.67 \cellcolor{gray!30}& 55.53 \cellcolor{gray!30}\\
        \quad LLaVA-1.5-7B & 20.50 & 47.53 & 27.10 & 42.00 & 60.61 & 47.80 & 31.25 \cellcolor{gray!30}& 54.07 \cellcolor{gray!30}& 37.45 \cellcolor{gray!30}\\
    
        \multicolumn{6}{l}{\small \textit{\textbf{Open Source Medical VLMs}}}\\
        
        \quad Med-LLaVa 7B & 2.10 & 37.56 & 31.25 & 2.35 & 13.89 & 39.25 & 2.23 \cellcolor{gray!30}& 25.73 \cellcolor{gray!30}& 35.25 \cellcolor{gray!30}\\
        \quad Med-Gemma 4B & 11.50 & 41.30 & 15.93 & 35.50 & 53.59 & 39.15 & 23.50 \cellcolor{gray!30}& 47.45 \cellcolor{gray!30}& 27.54 \cellcolor{gray!30}\\
        
    \bottomrule         
    \end{tabular}
    }
    \caption{Model Benchmarking for \textbf{English} with average scores across Chest X-Ray and Medi-CAT datasets.}
    \label{tab:model_benchmark_english}
\end{table*}

\section{BanglaMedVQA Data Analysis}
\subsection{Data Statistics}
The proposed dataset comprises a total of 2,000 VQA instances. The data are sourced equally from two prominent biomedical datasets, 1,000 instances from ChestX-ray14 and 1,000 from MedICaT. The dataset is balanced with respect to healthy and abnormal cases and is uniformly distributed across five question categories: \textit{modality}, \textit{organ}, \textit{abnormality}, \textit{condition}, and \textit{position}, each representing approximately 20\% of the total questions. Analysis of question keywords, as shown in Figure~\ref{fig:key_distribution}, illustrates a strong focus on imaging interpretation, anatomical localization, and abnormality assessment. The most frequent terms include \textit{abnormality}, \textit{condition}, \textit{organ}, \textit{technique}, and specific findings such as \textit{mass}, \textit{effusion}, \textit{atelectasis}, \textit{cardiomegaly}, \textit{nodule}, \textit{pneumonia}, and \textit{infiltrate}, reflecting the dataset's emphasis on clinical image understanding and diagnostic reasoning.

\subsection{Clinical Conditions}
The distribution of clinical conditions in the dataset is summarized in Figure~\ref{fig:condition_distribution}. Organ-wise, the majority of conditions target the \textit{Lungs}, followed by \textit{Heart \& Vessels}, \textit{Brain \& CNS}, \textit{Pleura}, \textit{Bones \& Spine}, \textit{Thorax}, \textit{Abdomen}, and \textit{Neoplastic} cases. Within each organ/system, the dataset covers a wide range of conditions, with more frequent labels including \textit{Infiltration}, \textit{Atelectasis}, \textit{Effusion}, \textit{Cardiomegaly}, \textit{Nodule}, \textit{Pneumothorax}, \textit{Mass}, \textit{Pneumonia}, \textit{Consolidation}, \textit{Pleural thickening}, \textit{Edema}, \textit{Emphysema}, \textit{Tumor}, and \textit{Lesion}, as well as rarer conditions such as \textit{Fibrosis}, \textit{Hemorrhage}, \textit{Calcification}, \textit{Fractures}, \textit{Cysts}, \textit{Metastasis}, and \textit{Stroke}.

\begin{figure*}[t]
    \centering
    \includegraphics[width=0.9\textwidth]{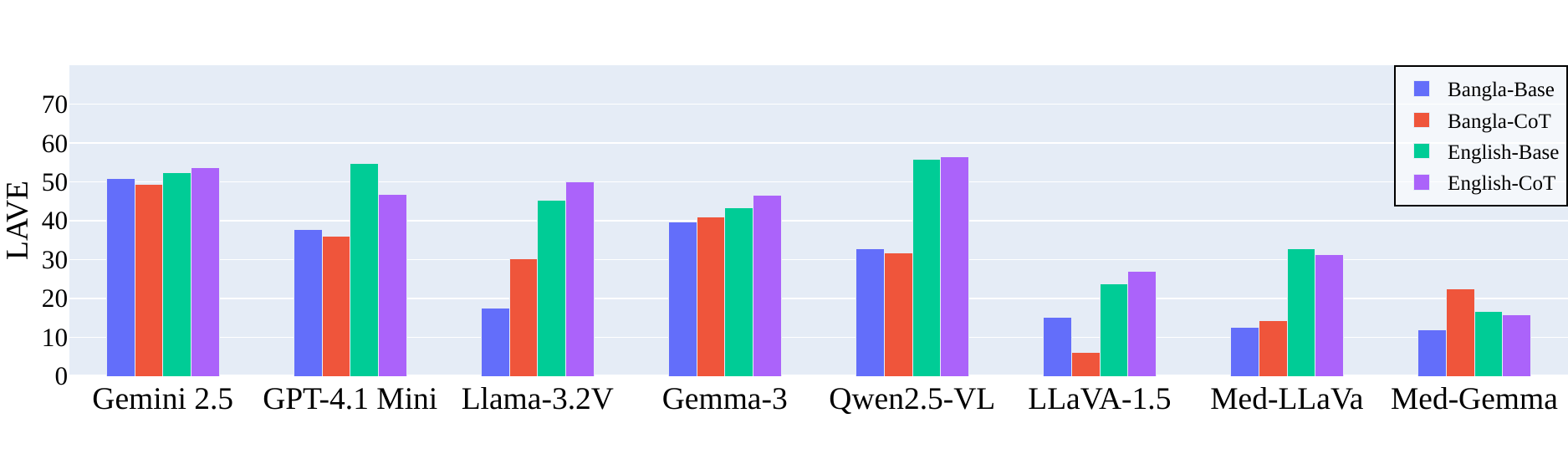} 
    \caption{LAVE score comparison of the Chest X-Ray dataset for different models under four different settings: vanilla (baseline performance) for Bangla, chain-of-thought (CoT) reasoning for Bangla, vanilla (baseline performance) for English, chain-of-thought (CoT) reasoning for English.}
    \label{fig:chest_xray}
\end{figure*}

\begin{figure*}[t]
    \centering
    \includegraphics[width=0.9\textwidth]{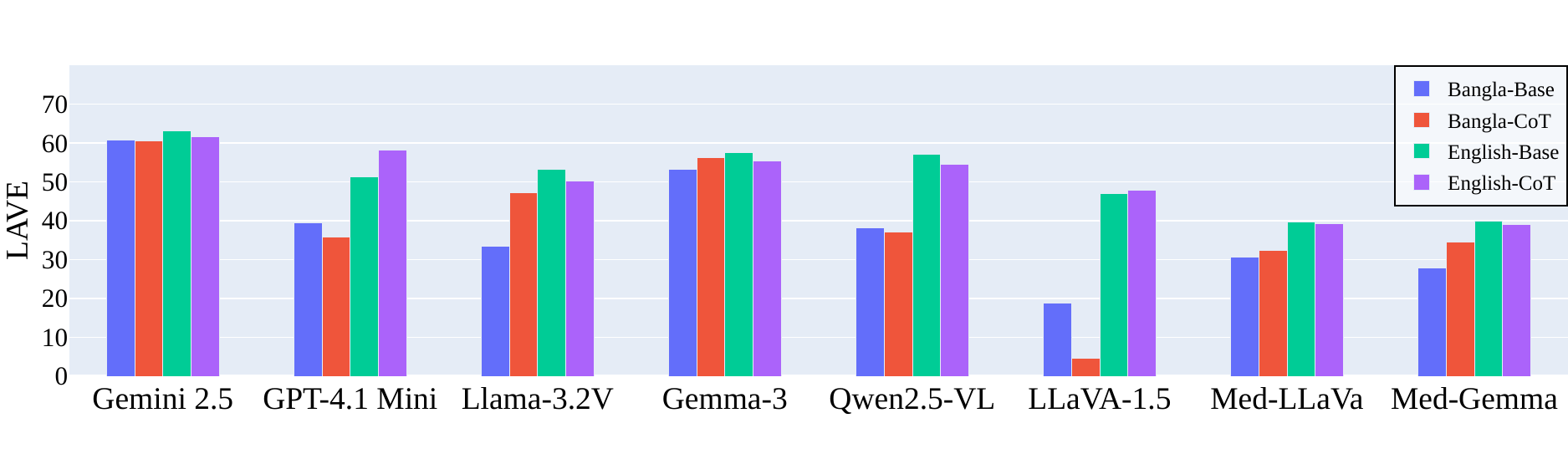} 
    \caption{LAVE score comparison of the MedICat dataset for different models under four different settings: vanilla (baseline performance) for Bangla, chain-of-thought (CoT) reasoning for Bangla, vanilla (baseline performance) for English, chain-of-thought (CoT) reasoning for English.}
    \label{fig:medicat}
\end{figure*}

\section{Experiment Setup}

\subsection{Experimentation with Bangla Question}
\noindent \textbf{Zero-Shot Prompting.}
To assess the medical reasoning capabilities of large vision-language models (LVLMs), we adopt a \textit{zero-shot prompting} approach. Each model receives a medical image $I$, a system prompt $P$, and a Bangla natural language question $Q_{\text{BAN}}$. Without any task-specific fine-tuning or example demonstrations, the model is expected to generate an open-ended response based solely on this input.

\noindent \textbf{Chain-of-Thought (CoT) Prompting.}
For \textit{Chain-of-Thought (CoT)} prompting, we explicitly guide the model to reason through the problem before producing an answer. This is done by appending the phrase \textbf{\textit{“Let's think step by step”}} to the system prompt $P$, encouraging multi-step reasoning and intermediate inference prior to answer generation.

\subsection{Experimentation with English}
To investigate the effect of language variation, we further extend our experiments to include English-language questions. Specifically, since the Bangla question–answer pairs were originally translated from English, we utilize their corresponding English versions $Q_{\text{ENG}}$ and $A_{\text{ENG}}$ for each medical image $I$ in this experiment to ensure consistency and enable cross-linguistic comparison.

For the English-based experiments, we maintain the same system prompt $P$ and input the English question $Q_{\text{ENG}}$ in place of the original Bangla question $Q_{\text{BAN}}$. This setup is used for all the experiments.

\subsection{Experimented Models}
We conduct a systematic evaluation of eight LVLMs, spanning closed-source, open-source, and domain-specific medical variants, on our dataset. Specifically, we benchmark two proprietary models, Gemini 2.5 Flash~\cite{gemini2025} and GPT-4.1 Mini~\cite{gpt4.1mini2024}, alongside a suite of open-source general-purpose models including LLaMA-3.2V~\cite{LLama-3.2}, Gemma-3~\cite{Gemma-3}, Qwen2.5-VL~\cite{Qwen-2.5}, and LLaVA-1.5~\cite{LLava-1.5}. In addition, we assessed two medical-domain LVLMs: Med-LLaVA~\cite{MedLLaVa} and Med-Gemma~\cite{medgemma}.

\subsection{Evaluation Metrics}
Performance of the models is reported using three complementary metrics: Accuracy (Acc), BERTScore (BScore), and LLM-Assisted VQA Evaluation(LAVE) \cite{manas2024improving}. Accuracy measures the proportion of exact matches between predicted and ground-truth answers. BERTScore evaluates the semantic similarity between predicted and reference answers using contextual embeddings. LAVE metric is designed to provide a more reliable evaluation of model performance in VQA tasks.


\section{Result and Analysis}

\begin{figure*}[t]
    \centering
    \hfill
    \begin{subfigure}[b]{0.45\textwidth}
    \includegraphics[width=\textwidth]{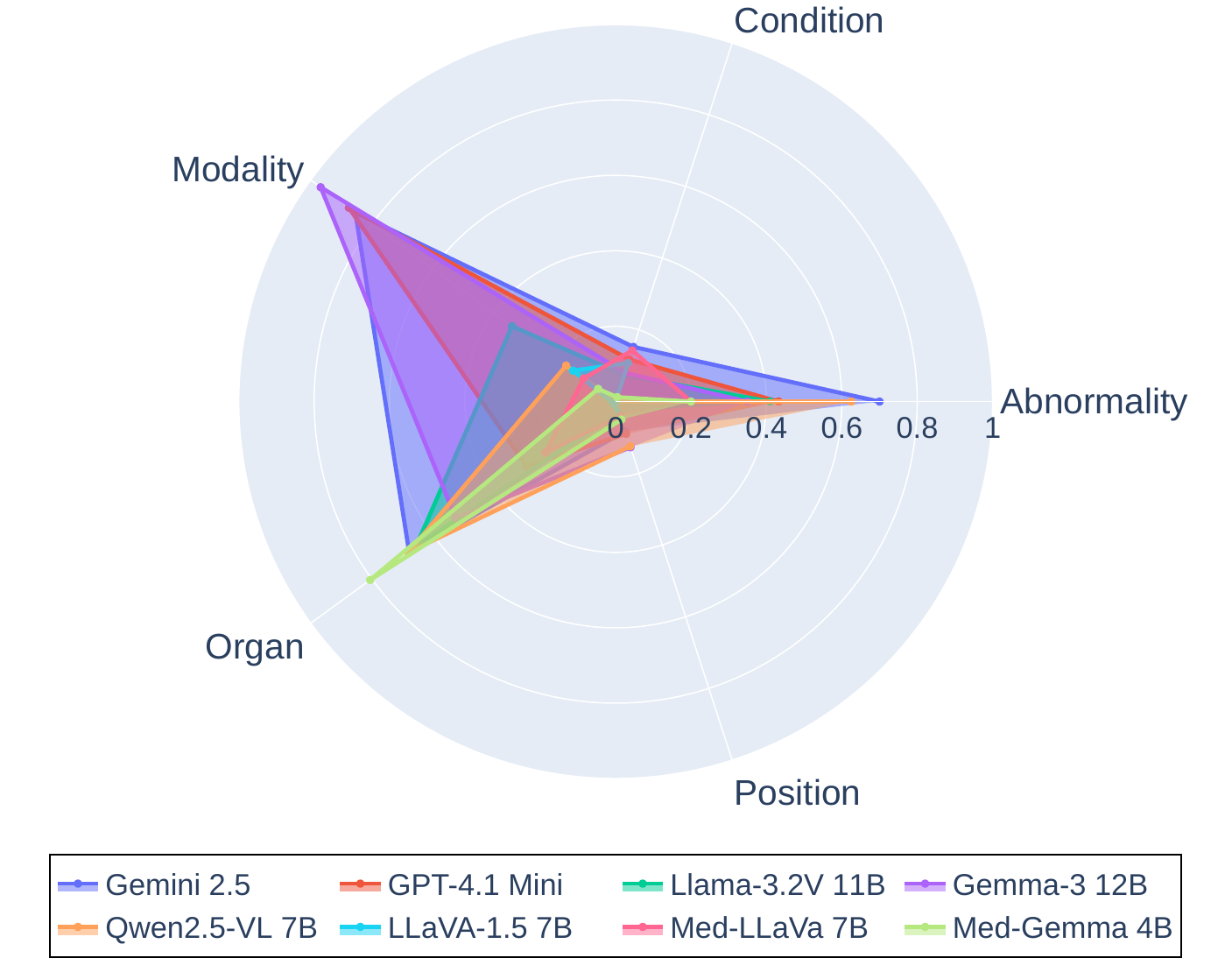}
    \caption{CoT prompt on Bangla QA pair}
    \label{fig:spider1}
    \end{subfigure}
    \hfill
    \begin{subfigure}[b]{0.45\textwidth}        
    \includegraphics[width=\textwidth]{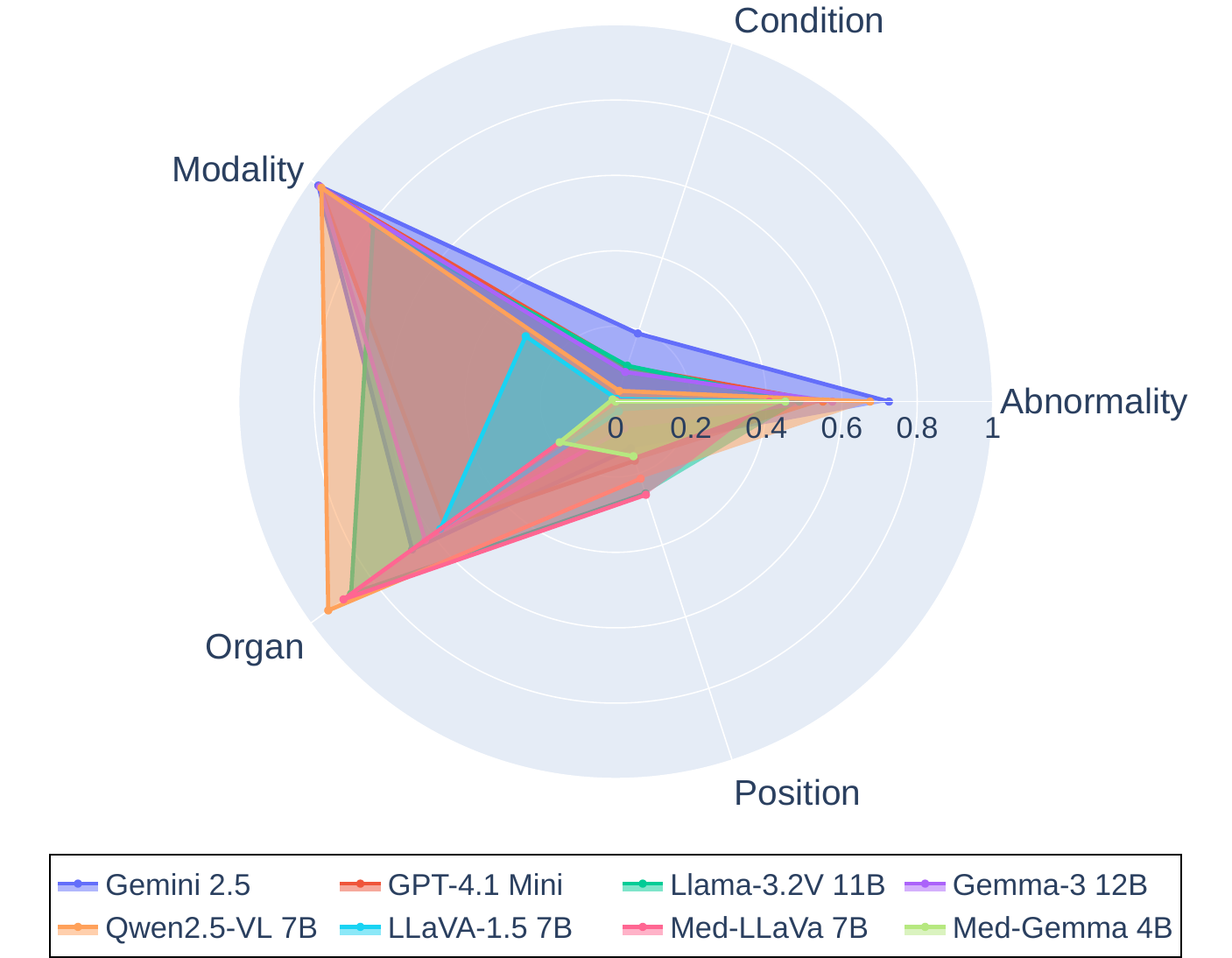}
    \caption{CoT prompt on English QA pair}
    \label{fig:spider2}
    \end{subfigure}
    \label{fig:chest_xray_specialized_ques}
    \caption{LAVE score comparison on the Chest X-Ray dataset across categorical question types with chain-of-thought reasoning in Bangla and English.}

\end{figure*}

\begin{figure*}[h]
    \centering
    \hfill
    \begin{subfigure}[b]{0.45\textwidth}
    \includegraphics[width=\textwidth]{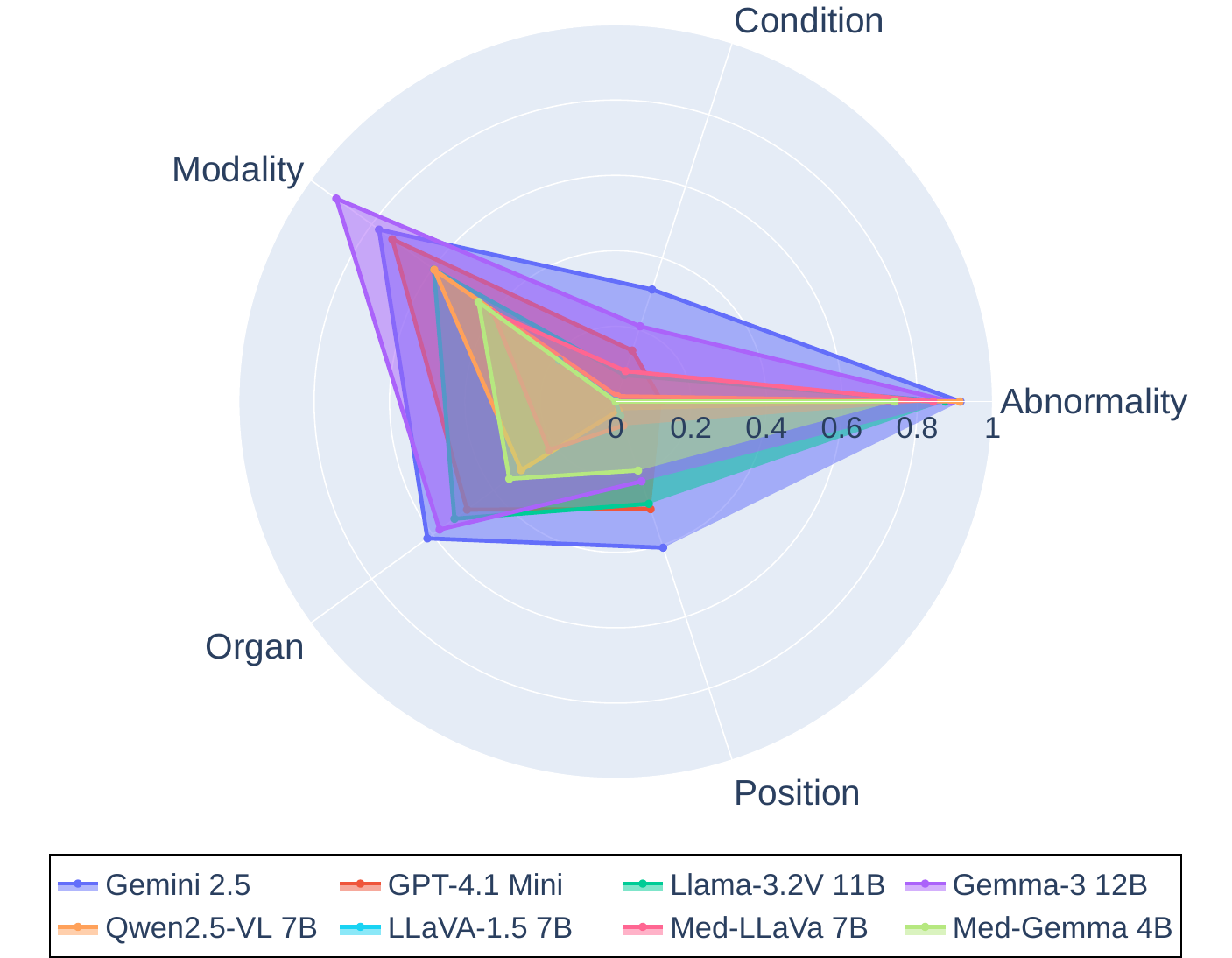}
    \caption{CoT prompt on Bangla QA pair}
    \label{fig:spider3}
    \end{subfigure}
    \hfill
    \begin{subfigure}[b]{0.45\textwidth}        
    \includegraphics[width=\textwidth]{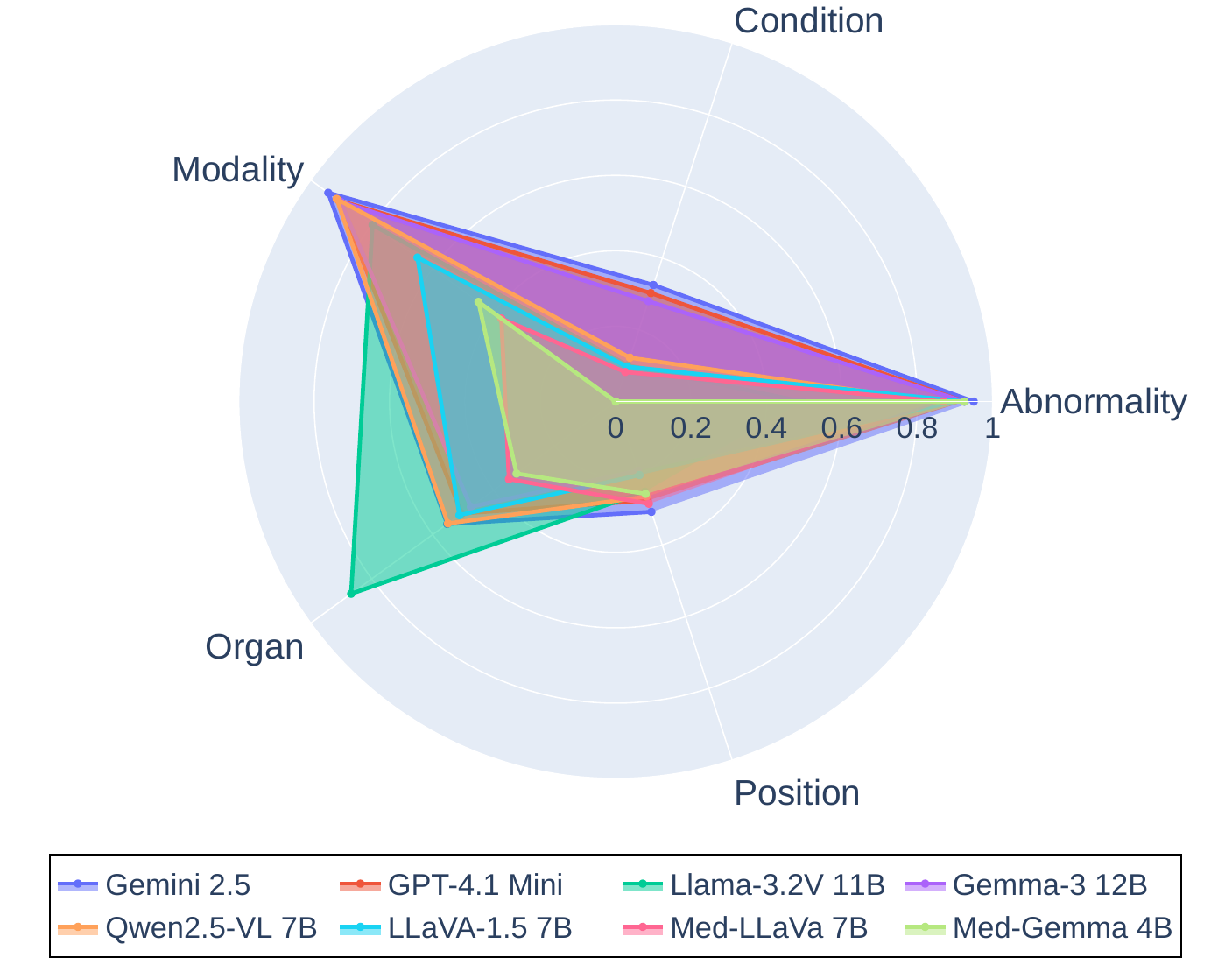}
    \caption{CoT prompt on English QA pair}
    \label{fig:spider4}
    \end{subfigure}
    \label{fig:medi_cat_specialized_ques}
    \caption{LAVE score comparison on the MedICat dataset across categorical question types with chain-of-thought reasoning in Bangla and English}
\end{figure*}

\begin{figure*}[h]
    \centering
    \includegraphics[width=0.98\textwidth]{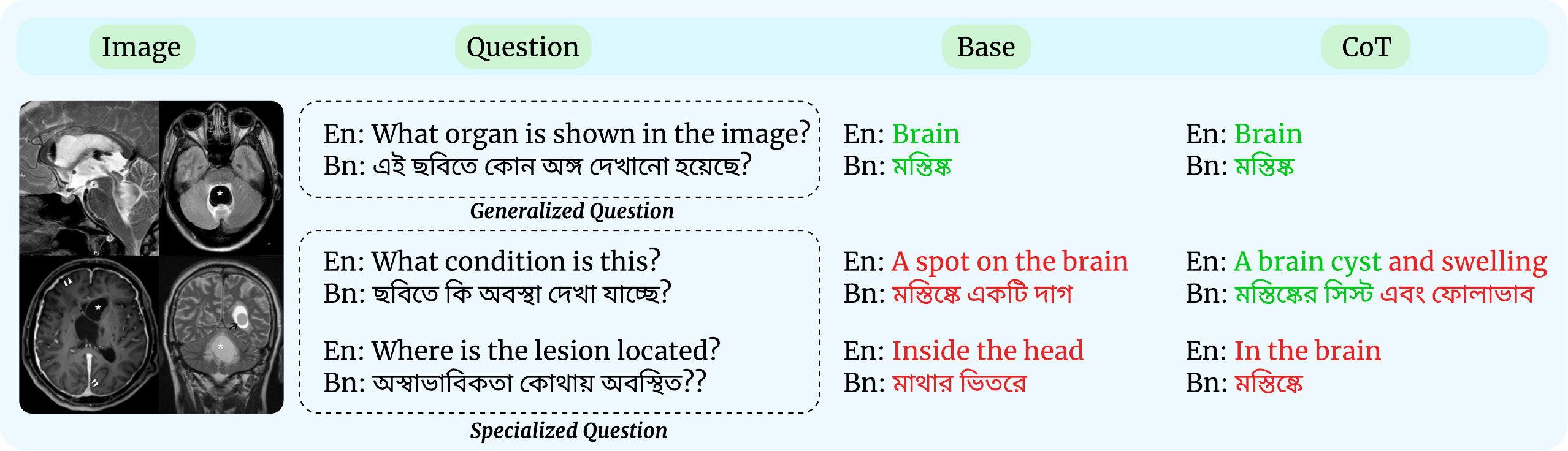} 
    \caption{Error Analysis of Medical Visual Question Answering (MedVQA) pairs from the proposed dataset, showcasing aligned English–Bangla questions and clinically validated answers across categories such as condition and abnormality localization.}
    \label{fig:error_analysis}
\end{figure*}

Results on Bangla and English QA pairs are presented in Tables~\ref{tab:model_benchmark} ,~\ref{tab:model_benchmark_english} and Figure ~\ref{fig:chest_xray}, ~\ref{fig:medicat} respectively.

\subsection{Bangla QA Pairs}
Overall, closed-source VLMs substantially outperform both open-source and medical-domain counterparts under zero-shot and chain-of-thought (CoT) prompting. Gemini 2.5 Flash achieves the highest LAVE scores in both Chest X-Ray and MedICaT, with 50.80\% and 60.08\%, respectively, under zero-shot prompting. In contrast, GPT-4.1 Mini performs significantly worse, with accuracy between 15--16\% and weaker LAVE alignment. Among open-source general-purpose VLMs, Gemma-3 12B achieves the strongest performance, reaching 39.70\% (ChestX-Ray) and 53.37\% (MedICaT) in zero-shot settings, with further gains under CoT prompting. Medical-specific VLMs such as Med-LLaVA and Med-Gemma underperform relative to general-purpose models, likely due to limited multilingual adaptation, with accuracies frequently dropping below 10\% in Bangla.  

\subsection{English QA Pairs}
In English, the performance gap between closed-source and open-source VLMs is narrower. Both Gemini 2.5 Flash and Qwen2.5-VL 7B yield competitive results, with Qwen2.5-VL achieving the best overall LAVE on Chest X-Ray (55.85\%) and Gemini 2.5 Flash leading on MedICaT (63.15\%) under zero-shot prompting. GPT-4.1 Mini, while weak in Bangla, demonstrates stronger alignment in English, with accuracy exceeding 40\% on Chest X-Ray. Interestingly, the open-source Gemma-3 12B performs competitively, surpassing GPT-4.1 Mini on MedICaT. Medical-domain VLMs again fail to generalize, with Med-LLaVA yielding near-zero accuracy across both datasets.  

\subsection{Impact of Chain-of-Thought Prompting}
We observe a consistent trend where CoT prompting improves the performance of most models, particularly open-source VLMs. For instance, Llama-3.2V 11B shows a notable increase in accuracy and LAVE, from 17.50\% to 30.17\% in Bangla and from 45.30\% to 50.12\% in English. Closed-source Gemini 2.5 Flash also benefits, particularly in Bangla, though the gains are more modest. In contrast, medical VLMs such as Med-LLaVA generally show negligible gains, reflecting their limited reasoning in multilingual contexts. Med-Gemma, however, demonstrates substantial improvements under CoT prompting, highlighting its potential when guided by reasoning-based prompts.

\subsection{Performance Across Specialized Questions}
We further evaluate model performance across different question categories. The results are presented in Figures~\ref{fig:spider1}--\ref{fig:spider4} and Tables~\ref{tab:chest_dataset_performance_overall_all_category} and \ref{tab:medi_cat_dataset_performance_overall_all_category}.The questions can be broadly divided into two categories: \textit{general} questions, which involve tasks such as identifying image modality or organ type, and \textit{specialized} questions, which require deeper medical knowledge to assess conditions and anatomical positions.

While closed-source models such as Gemini-2.5 and GPT-4.1 Mini outperform other models and achieve relatively strong results on general questions, their performance on specialized questions remains severely limited. For Bangla, under chain-of-thought reasoning across the entire dataset, Gemini-2.5 achieves the highest average specialized LAVE score, but only 23.25\% on \textit{Condition} and 24.50\% on \textit{Position}. GPT-4.1 Mini performs even worse, with scores of 13\% and 19.50\% on these categories, respectively. These results indicate that even the strongest models perform close to random guessing on specialized diagnostic tasks.  

Among open-source models, Gemma-3 consistently outperforms other open-source counterparts across all categories and occasionally surpasses closed-source models such as GPT on general questions. However, it struggles significantly on specialized questions. Furthermore, the performance of open-source medical VLMs on these specialized categories is alarmingly low, highlighting a substantial gap in their ability to support real-world medical diagnosis.

\subsection{Cross-Lingual Observations.}
Comparing Bangla and English benchmarks, we find that performance in Bangla is generally weaker, particularly for open-source and domain-specific VLMs. Closed-source VLMs like Gemini exhibit stronger multilingual transfer, maintaining relatively robust scores in Bangla. This suggests that large-scale multilingual pretraining and broader instruction tuning are critical for achieving cross-lingual generalization in medical VQA. Our findings highlight three key insights: (1) closed-source VLMs maintain superior robustness and generalization across languages, (2) CoT prompting consistently enhances reasoning and alignment, and (3) medical-domain VLMs, while specialized, show limited cross-lingual capability and require further adaptation for multilingual medical tasks.

\subsection{Error Analysis}
In Fig \ref{fig:error_analysis}, we also exhibit qualitative analysis. The result shows that while the models generally handle straightforward questions such as identifying the imaging modality or the organ involved successfully, they struggle with more specialized diagnostic reasoning. Errors are most frequent in questions requiring precise localization or interpretation of pathological findings, where models often confuse nearby anatomical regions or misclassify related conditions such as cysts and edema. In several cases, responses were partially correct, capturing general abnormalities but missing more precise clinical details. These limitations highlight gaps in visual–textual grounding and domain understanding.

\section{Conclusion}
We are the first to propose Medical Visual Question Answering for Bangla, addressing a significant gap in low-resource language evaluation for multimodal AI systems. Through systematic evaluation of both open-source and closed-source LLMs and LVLMs, we observed that while top-performing models such as Gemini and GPT-4.1 mini achieve reasonable performance on general questions, they struggle severely on specialized diagnostic tasks, often performing close to random. Open-source models, including Gemma-3, occasionally outperform closed-source models on general questions but also fail on specialized medical queries. Incorporating chain-of-thought prompting provides moderate improvements, highlighting insufficient visual understanding as a key limitation. The BanglaMedVQA dataset thus serves as a crucial benchmark to drive future progress in this underserved area.   

\section*{Limitations}
Despite the contributions of this work, several limitations remain. While our evaluation provides valuable insights into the performance of existing models, we were unable to improve diagnostic accuracy through model fine-tuning. Fine-tuning large multimodal models on our BanglaMedVQA dataset could potentially yield significant performance gains; however, due to computational and resource constraints, this was not feasible within the scope of this study.

\section*{Ethical Statement}
This work involves the creation and use of a medical visual question-answering dataset in Bangla. All images and associated annotations were sourced from publicly available datasets or generated with appropriate institutional permissions. Annotations and dataset curation were performed by qualified medical experts to ensure accuracy and minimize the risk of misinformation. We acknowledge that models trained on this dataset are intended for research purposes only and should not be used for clinical decision-making. Patient data privacy has been maintained throughout, and no personally identifiable information is included in the dataset. We encourage responsible use of both the dataset and models, and recommend thorough evaluation before any deployment in real-world medical contexts.

\bibliography{aaai2026}
\appendix
\section{Question Category Wise Performance Comparison}

Tables~\ref{tab:chest_dataset_performance_overall_all_category} and~\ref{tab:medi_cat_dataset_performance_overall_all_category} present category-wise results for all evaluated models under two experimental settings, Base, Chain-of-Thought (CoT) for the \textit{Chest X-Ray} and \textit{MedICaT} datasets. The questions are divided into \textit{general} categories (\textit{Modality}, \textit{Organ}) and \textit{specialized} categories (\textit{Abnormality}, \textit{Condition}, \textit{Position}).

\subsubsection{Bangla QA Results}

In the Bangla benchmark, closed-source models dominate under the base prompt. Gemini 2.5 achieves an average of 87.50\% on Modality and 80.75\% on Abnormality, substantially outperforming open-source models in general questions. GPT-4.1 Mini performs slightly lower, with 85.25\% on Modality and 47.50\% on Abnormality, indicating a consistent gap between the two closed-source systems. Accuracy drops sharply for specialized categories, with Condition and Position reaching only 22.75\% and 22.87\%, respectively. This trend demonstrates that even leading closed-source models struggle with relational and spatial reasoning required for specialized medical questions in Bangla.

Among open-source VLMs, Gemma-3 12B achieves 93.70\% on Modality and 30.00\% on Abnormality, while Qwen2.5-VL 7B scores 13.00\% on Modality but 72.50\% on Abnormality, highlighting that different open-source architectures show partial strengths in specific general categories, yet consistently underperform in reasoning-heavy specialized categories such as Condition (8.00\%–2.85\%) and Position (12.75\%–15.25\%). Open-source medical VLMs show similarly constrained performance. Med-LLaVa 7B reaches 6.75\% for both Modality and Organ, while achieving 39.75\% on Abnormality, and Med-Gemma 4B performs slightly better on Organ (50.00\%) but fails entirely on certain specialized questions (Abnormality 0.00\%, Condition 0.50\%, Position 3.00\%). These results suggest that medical pretraining alone does not compensate for the challenges of Bangla language understanding or reasoning-intensive QA.

Under CoT prompting, closed-source models exhibit modest improvements across most categories. Gemini 2.5 shows an increase to 86.125\% for Abnormality, 27.25\% for Condition, and 32.625\% for Position, indicating that stepwise reasoning partially aids certain visual–clinical inferences. GPT-4.1 Mini improves slightly as well, yet remains significantly behind in specialized reasoning categories. Open-source models also gain from CoT; Gemma-3 12B improves to 75.50\% on Abnormality, Qwen2.5-VL 7B rises to 84.06\% on the same category, and Llama-3.2V 11B reaches 64.25\% on Abnormality. Despite these gains, specialized reasoning for Condition and Position remains consistently weak, with almost all open-source and medical VLMs scoring below 20\%.

\subsubsection{English QA Results}

In the English benchmark, models demonstrate notably stronger performance across all categories compared to the Bangla setting, yet they continue to struggle in specialized clinical reasoning tasks. Closed-source models again lead the benchmark, with Gemini 2.5 and GPT-4.1 Mini showing consistent superiority under the base prompt. Gemini 2.5 attains 97.0\% on Modality and 72.25\% on Abnormality, while GPT-4.1 Mini slightly surpasses it on Abnormality with 75.0\%. Despite their high accuracy in general categories, performance drops sharply for specialized ones Condition (21.75\%) and Position (19.25\%) indicating that even in English, reasoning-heavy and spatially dependent questions remain challenging.

Open-source models also exhibit substantial improvement in English, with Qwen2.5-VL 7B achieving near closed-source performance 97.0\% on Modality and 94.25\% on Organ demonstrating strong multimodal grounding in English. Similarly, Llama-3.2V 11B and Gemma-3 12B show balanced outcomes across general questions. However, their accuracy still collapses below 30\% in relationally complex tasks such as Condition and Position, reaffirming that the bottleneck lies in fine-grained visual–textual reasoning rather than linguistic comprehension.

Under CoT prompting, all models particularly open-source ones exhibit noticeable gains. Qwen2.5-VL 7B reaches an overall accuracy of 56.55\%, the highest among open models, even surpassing some closed systems. Gemini 2.5 maintains top-tier results with 97.5\% on Modality and 72.5\% on Abnormality, suggesting robustness to prompt variation. Nonetheless, specialized categories show only marginal improvements, with accuracy in Condition and Position remaining below 30\% across nearly all models. Medical-domain-tuned systems such as Med-LLaVA 7B and Med-Gemma 4B continue to lag, indicating that domain specialization without strong reasoning or cross-lingual adaptation provides limited benefit.

\begin{table*}[t]
    \centering
    \small
    \resizebox{.8\textwidth}{!}{
    \begin{tabular}{lcccccc}
        \toprule
        \multirow{2}{*}{\textbf{Models}} & \multicolumn{2}{c}{\textbf{Generalized Question}} 
        & \multicolumn{3}{c}{\textbf{Specialized Question}} & \multirow{2}{*}{\textbf{Overall}} \\
        \cmidrule(lr){2-3}
        \cmidrule(lr){4-6}
        & \textbf{Modality} & \textbf{Organ} & \textbf{Abnormality} & \textbf{Condition} & \textbf{Position} & \\    
        \midrule
        \multicolumn{7}{c}{\small \textit{\textbf{Base Prompt on Bangla Chest X-Ray QA Pair}}}\\
        \midrule
        \multicolumn{7}{l}{\small \textit{\textbf{Closed Source VLMs}}}\\
        \quad Gemini 2.5     & 95.75 & \textbf{65.00} & 70.00 & \textbf{15.75} & 7.50  & \textbf{50.80} \cellcolor{gray!30}\\
        \quad GPT-4.1 Mini   & 85.25 & 29.75 & 47.50 & 11.00 & 15.00 & 37.70 \cellcolor{gray!30}\\
        \midrule
        \multicolumn{7}{l}{\small \textit{\textbf{Open Source VLMs}}}\\
        \quad Llama-3.2V 11B & 11.00 & 29.50 & 40.00 & 3.00  & 1.75 & 17.50 \cellcolor{gray!30}\\
        \quad Gemma-3 12B    & \textbf{97.50} & 50.25 & 30.00 & 8.00  & 12.75 & 39.70 \cellcolor{gray!30}\\
        \quad Qwen2.5-VL 7B  & 13.00 & 63.50 & \textbf{72.50} & 2.85  & \textbf{15.25} & 32.85 \cellcolor{gray!30}\\
        \quad LLaVA-1.5 7B   & 13.50 & 2.00  & 42.50 & 14.75 & 3.25  & 15.20 \cellcolor{gray!30}\\
        \midrule
        \multicolumn{7}{l}{\small \textit{\textbf{Open Source Medical VLMs}}}\\
        \quad Med-LLaVa 7B   & 6.75  & 6.75  & 39.75 & 8.00  & 2.25  & 12.70 \cellcolor{gray!30}\\
        \quad Med-Gemma 4B   & 6.00  & 50.00 & 0.00  & 0.50  & 3.00  & 11.90 \cellcolor{gray!30}\\ 
        \midrule
        \multicolumn{7}{c}{\small \textit{\textbf{Base Prompt on English Chest X-Ray QA Pair}}}\\
        \midrule
        \multicolumn{7}{l}{\small \textit{\textbf{Closed Source VLMs}}}\\
        \quad Gemini 2.5     & \textbf{97.00} & 65.25 & 72.25 & 18.58 & 9.25 & \cellcolor{gray!30} \textbf{52.47} \\
        \quad GPT-4.1 Mini   & 97.50 & 60.00 & \textbf{75.00} & \textbf{21.75} & 19.25 & \cellcolor{gray!30}54.70 \\
        \midrule
        \multicolumn{7}{l}{\small \textit{\textbf{Open Source VLMs}}}\\
        \quad Llama-3.2V 11B & 67.25 & 67.75 & 67.50 & 10.00 & 14.00 & \cellcolor{gray!30}45.30 \\
        \quad Gemma-3 12B    & 76.25 & 66.25 & 47.50 & 11.50 & 14.00 & \cellcolor{gray!30}43.30 \\
        \quad Qwen2.5-VL 7B  & 97.00 & \textbf{94.25} & 60.00 & 2.25  & \textbf{25.75} & \cellcolor{gray!30}55.85 \\
        \quad LLaVA-1.5 7B   & 8.25  & 59.50 & 45.00 & 4.75  & 2.00  & \cellcolor{gray!30}23.90 \\
        \midrule
        \multicolumn{7}{l}{\small \textit{\textbf{Open Source Medical VLMs}}}\\
        \quad Med-LLaVa 7B   & 3.00  & 88.25 & 40.25 & 8.75  & 23.50 & \cellcolor{gray!30}32.75 \\
        \quad Med-Gemma 4B   & 0.00  & 23.25 & 45.00 & 0.00  & 15.25 & \cellcolor{gray!30}16.70 \\  
        \midrule
        \multicolumn{7}{c}{\small \textit{\textbf{CoT Prompt on Bangla Chest X-Ray QA Pair}}}\\
        \midrule
        \multicolumn{7}{l}{\small \textit{\textbf{Closed Source VLMs}}}\\
        \quad Gemini 2.5      & 81.51 & \textbf{64.75} & \textbf{80.75} & \textbf{23.25} & \textbf{24.50} & \cellcolor{gray!30} \textbf{49.33} \\
        \quad GPT-4.1 Mini    & 80.38 & 39.00 & 27.75 & 13.00 & 19.50 & \cellcolor{gray!30}36.15 \\
        \midrule
        \multicolumn{7}{l}{\small \textit{\textbf{Open Source VLMs}}}\\
        \quad Llama-3.2V 11B  & 46.81 & 59.19 & 64.25 & 7.39  & 15.75 & \cellcolor{gray!30}30.17 \\
        \quad Gemma-3 12B     & \textbf{94.19} & 55.13 & 62.00 & 14.50 & 17.50 & \cellcolor{gray!30}41.00 \\
        \quad Qwen2.5-VL 7B   & 37.88 & 49.50 & 76.88 & 0.75  & 7.50  & \cellcolor{gray!30}31.85 \\
        \quad LLaVA-1.5 7B    & 16.25 & 0.38  & 0.00  & 5.50  & 4.38  & \cellcolor{gray!30}6.05 \\
        \midrule
        \multicolumn{7}{l}{\small \textit{\textbf{Open Source Medical VLMs}}}\\
        \quad Med-LLaVa 7B    & 25.38 & 22.44 & 52.13 & 11.38 & 5.35  & \cellcolor{gray!30}14.35 \\
        \quad Med-Gemma 4B    & 25.38 & 57.69 & 47.00 & 0.63  & 12.13 & \cellcolor{gray!30}22.50 \\   
       \midrule
        \multicolumn{7}{c}{\small \textit{\textbf{CoT Prompt on English Chest X-Ray QA Pair}}}\\
        \midrule
        \multicolumn{7}{l}{\small \textit{\textbf{Closed Source VLMs}}}\\
        \quad Gemini 2.5      & \textbf{97.50} & 66.75 & \textbf{72.50} & \textbf{19.00} & 13.00 & \cellcolor{gray!30}53.75 \\
        \quad GPT-4.1 Mini    & 97.00 & 56.00 & 55.00 & 9.75 & 16.50 & \cellcolor{gray!30}46.85 \\
        \midrule
        \multicolumn{7}{l}{\small \textit{\textbf{Open Source VLMs}}}\\
        \quad Llama-3.2V 11B  & 79.50 & 86.75 & 48.75 & 10.00 & \textbf{25.63} & \cellcolor{gray!30}50.12 \\
        \quad Gemma-3 12B     & 97.00 & 62.50 & 57.50 & 8.25 & 7.50 & \cellcolor{gray!30}46.55 \\
        \quad Qwen2.5-VL 7B   & 96.50 & \textbf{94.25} & 67.50 & 3.00 & 21.50 & \cellcolor{gray!30}\textbf{56.55} \\
        \quad LLaVA-1.5 7B    & 29.50 & 57.50 & 45.00 & 0.75 & 2.75 & \cellcolor{gray!30}27.10 \\
        \midrule
        \multicolumn{7}{l}{\small \textit{\textbf{Open Source Medical VLMs}}}\\
        \quad Med-LLaVa 7B    & 0.00 & 89.25 & 40.75 & 0.25 & 26.00 & \cellcolor{gray!30}31.25 \\
        \quad Med-Gemma 4B    & 1.00 & 18.38 & 45.00 & 0.00 & 15.25 & \cellcolor{gray!30}15.93 \\
        \bottomrule
      
    \end{tabular}
    }
    \caption{Model performance for different categorical questions on the Chest X-ray Dataset}
    \label{tab:chest_dataset_performance_overall_all_category}
\end{table*}

\begin{table*}[t]
    \centering
    \small
    \resizebox{.8\textwidth}{!}{
    \begin{tabular}{lcccccc}
        \toprule
        \multirow{2}{*}{\textbf{Models}} & \multicolumn{2}{c}{\textbf{Generalized Question}} 
        & \multicolumn{3}{c}{\textbf{Specialized Question}} & \multirow{2}{*}{\textbf{Overall}} \\
        \cmidrule(lr){2-3}
        \cmidrule(lr){4-6}
        & \textbf{Modality} & \textbf{Organ} & \textbf{Abnormality} & \textbf{Condition} & \textbf{Position} & \\    
        \midrule
        \multicolumn{7}{c}{\small \textit{\textbf{Base Prompt on Bangla MediCat QA Pair}}}\\
        \midrule
        \multicolumn{7}{l}{\small \textit{\textbf{Closed Source VLMs}}}\\
        \quad Gemini 2.5     & 79.25 & \textbf{61.63} & \textbf{91.50} & \textbf{29.75} & \textbf{38.25} & \cellcolor{gray!30}\textbf{60.08} \\
        \quad GPT-4.1 Mini   & 79.63 & 47.38 & 8.50  & 23.75 & 38.25 & \cellcolor{gray!30}39.50 \\
        \midrule
        \multicolumn{7}{l}{\small \textit{\textbf{Open Source VLMs}}}\\
        \quad Llama-3.2V 11B & 44.00 & 32.25 & 83.00 & 1.75  & 6.50  & \cellcolor{gray!30}33.50 \\
        \quad Gemma-3 12B    & \textbf{89.88} & 57.75 & 83.75 & 14.75 & 20.75 & \cellcolor{gray!30}53.70 \\
        \quad Qwen2.5-VL 7B  & 64.50 & 30.00 & 89.00 & 4.25  & 3.75  & \cellcolor{gray!30}38.30 \\
        \quad LLaVA-1.5 7B   & 15.75 & 0.00  & 76.75 & 0.50  & 0.75  & \cellcolor{gray!30}18.75 \\
        \midrule
        \multicolumn{7}{l}{\small \textit{\textbf{Open Source Medical VLMs}}}\\
        \quad Med-LLaVa 7B   & 42.25 & 15.38 & 76.38 & 6.38  & 13.25 & \cellcolor{gray!30}30.72 \\
        \quad Med-Gemma 4B   & 36.13 & 16.25 & 67.13 & 0.00  & 19.50 & \cellcolor{gray!30}27.80 \\
        
        \midrule
        \multicolumn{7}{c}{\small \textit{\textbf{Base Prompt on English MediCat QA Pair}}}\\
        \midrule
        \multicolumn{7}{l}{\small \textit{\textbf{Closed Source VLMs}}}\\
        \quad Gemini 2.5     & \textbf{93.50} & 57.25 & \textbf{95.00} & \textbf{30.50} & \textbf{39.50} & \cellcolor{gray!30}\textbf{63.15} \\
        \quad GPT-4.1 Mini   & 89.00 & 50.00 & 60.00 & 21.50 & 36.08 & \cellcolor{gray!30}51.31 \\
        \midrule
        \multicolumn{7}{l}{\small \textit{\textbf{Open Source VLMs}}}\\
        \quad Llama-3.2V 11B & 89.25 & 55.25 & 86.75 & 15.00 & 20.00 & \cellcolor{gray!30}53.25 \\
        \quad Gemma-3 12B    & 91.75 & 55.00 & 90.00 & 25.75 & 25.50 & \cellcolor{gray!30}57.60 \\
        \quad Qwen2.5-VL 7B  & 90.50 & \textbf{65.00} & 87.50 & 17.75 & 24.75 & \cellcolor{gray!30}57.10 \\
        \quad LLaVA-1.5 7B   & 55.00 & 52.50 & 92.50 & 16.25 & 18.75 & \cellcolor{gray!30}47.00 \\
        \midrule
        \multicolumn{7}{l}{\small \textit{\textbf{Open Source Medical VLMs}}}\\
        \quad Med-LLaVa 7B   & 45.00 & 35.00 & 89.25 & 3.25  & 26.25 & \cellcolor{gray!30}39.75 \\
        \quad Med-Gemma 4B   & 45.00 & 32.50 & 92.50 & 0.00  & 30.00 & \cellcolor{gray!30}40.00 \\    
        \midrule
        \multicolumn{7}{c}{\small \textit{\textbf{CoT Prompt on Bangla MediCat QA Pair}}}\\
        \midrule
        \multicolumn{7}{l}{\small \textit{\textbf{Closed Source VLMs}}}\\
        \quad Gemini 2.5      & 77.63 & \textbf{61.75} & \textbf{91.50} & \textbf{31.25} & \textbf{40.75} & \cellcolor{gray!30}\textbf{60.58} \\
        \quad GPT-4.1 Mini    & 73.25 & 48.75 & 12.25 & 14.25 & 30.00 & \cellcolor{gray!30}35.70 \\
        \midrule
        \multicolumn{7}{l}{\small \textit{\textbf{Open Source VLMs}}}\\
        \quad Llama-3.2V 11B  & 59.63 & 52.88 & 87.50 & 7.40  & 28.50 & \cellcolor{gray!30}47.18 \\
        \quad Gemma-3 12B     & \textbf{91.63} & 57.75 & 89.00 & 21.00 & 22.25 & \cellcolor{gray!30}56.33 \\
        \quad Qwen2.5-VL 7B   & 59.50 & 31.00 & 91.25 & 1.50  & 2.50  & \cellcolor{gray!30}37.15 \\
        \quad LLaVA-1.5 7B    & 18.50 & 0.00  & 0.00  & 0.25  & 4.00  & \cellcolor{gray!30}4.55 \\
        \midrule
        \multicolumn{7}{l}{\small \textit{\textbf{Open Source Medical VLMs}}}\\
        \quad Med-LLaVa 7B    & 40.25 & 21.88 & 84.25 & 8.50  & 6.70  & \cellcolor{gray!30}32.31 \\
        \quad Med-Gemma 4B    & 45.00 & 34.88 & 74.00 & 0.00  & 19.25 & \cellcolor{gray!30}34.63 \\    
        \midrule
        \multicolumn{7}{c}{\small \textit{\textbf{CoT Prompt on English MediCat QA Pair}}}\\
        \midrule
        \multicolumn{7}{l}{\small \textit{\textbf{Closed Source VLMs}}}\\
        \quad Gemini 2.5      & \textbf{94.25} & 55.25 & \textbf{95.00} & \textbf{32.50} & \textbf{30.75} & \textbf{61.55} \cellcolor{gray!30}\\
        \quad GPT-4.1 Mini    & 90.75 & 50.50 & 92.50 & 30.25 & 27.58 & 58.31 \cellcolor{gray!30}\\
        \midrule
        \multicolumn{7}{l}{\small \textit{\textbf{Open Source VLMs}}}\\
        \quad Llama-3.2V 11B  & 79.75 & \textbf{86.75} & 49.50 & 10.13 & 25.13 & 50.25 \cellcolor{gray!30}\\
        \quad Gemma-3 12B     & 90.25 & 47.50 & 92.50 & 28.00 & 19.25 & 55.50 \cellcolor{gray!30}\\
        \quad Qwen2.5-VL 7B   & 91.50 & 55.00 & 85.00 & 12.25 & 26.50 & 54.50 \cellcolor{gray!30}\\
        \quad LLaVA-1.5 7B    & 65.00 & 51.25 & 92.50 & 9.75  & 20.50 & 47.80 \cellcolor{gray!30}\\
        \midrule
        \multicolumn{7}{l}{\small \textit{\textbf{Open Source Medical VLMs}}}\\
        \quad Med-LLaVa 7B    & 37.50 & 35.00 & 87.00 & 8.25  & 28.50 & 39.25 \cellcolor{gray!30}\\
        \quad Med-Gemma 4B    & 45.00 & 32.50 & 92.50 & 0.00  & 25.75 & 39.15 \cellcolor{gray!30}\\
        \bottomrule         
    \end{tabular}
    }
    \caption{Model performance for different categorical questions on MedICaT dataset}
    \label{tab:medi_cat_dataset_performance_overall_all_category}
\end{table*}


\section{Used Prompts in the Paper} \label{all_prompts}


\noindent \textbf{VQA Creation Prompt}
\begin{center}
\begin{tcolorbox}[colback=gray!5!white, colframe=black, title={Structured Prompt for Med-VQA Generation}, width=0.49\textwidth]
\small
You are a clinically grounded medical vision-language assistant. Your task is to create structured question–answer (QA) pairs from a given medical image and its metadata.\\[4pt]

\textbf{We Provide:}
\begin{itemize}
\item A medical image
\item Structured metadata describing that image
\end{itemize}

\textbf{Metadata Example:}
\texttt{
\{
  "modality": "X-ray",
  "organ": "Lung",
  "abnormality": ["Nodule"],
  "condition": ["Malignancy"],
  "position": ["Upper lobe of the right lung"]
\}
}\\[6pt] 

\textbf{Task Instructions:}
\begin{enumerate}
\item For each category --- modality, organ, abnormality, condition --- generate one structured question. Keep each answer as short as possible.
\item For the position category, generate one or more questions depending on the data in the metadata.
\item For the condition category:
\begin{itemize}
\item Generate a single question such as: ``What specific conditions are identified here?''
\item Include all conditions listed in the metadata in the answer, along with their positions if available.
\end{itemize}
\item For position, generate QA about the anatomical location of each disease.
\item Derive all answers only from the metadata and the image.
\item Do not invent findings that are not present in the metadata.
\end{enumerate}

\textbf{Output Format:}\\
Return JSON in the following structure:
\texttt{
\{
  "image\_id": "IMG\_001",
  "qa\_pairs": [
    \{
      "category": "modality",
      "question": "What is the imaging modality?",
      "answer": "X-ray"
    \},
    \{
      "category": "organ",
      "question": "Which organ is the focus of this image?",
      "answer": "Lung"
    \},
    \dots 
  ]
\}
} 

\end{tcolorbox}
\end{center}

\newpage
\noindent \textbf{Prompt Used for LAVE Evaluation}
\begin{center}
\begin{tcolorbox}[colback=gray!5!white, colframe=black, title={Prompt for LAVE evaluation}, width=0.49\textwidth]
\small
Acts as a judge to compute LAVE scores for reference/prediction pairs.\\
Compare each prediction with its reference. Output \textbf{ONLY} a JSON list of floats (0 to 1). No extra text.\\[4pt]
"You are a strict evaluator."\\
"Return ONLY a valid JSON array of floats between 0 and 1."\\
"Each float represents similarity between reference and prediction."\\
"If a reference answer is missing, treat it as 'No'."\\
"Do NOT return any text outside the JSON."
\end{tcolorbox}
\end{center}

\noindent \textbf{ZeroShot Prompt}
\begin{center}
\begin{tcolorbox}[colback=gray!5!white, colframe=black, title={ZeroShot Prompt for Bangla Med-VQA}, width=0.49\textwidth]
\small
You are a careful, clinically grounded medical vision-language assistant.\\
\textbf{Task:} You will be given a medical image and a single question about that image (e.g., modality, organ, abnormality, condition, or the position of the condition). Your job is to look at the image and provide the exact, minimal answer to the question.\\[4pt]

\textbf{CRITICAL:} You must respond with \textbf{ONLY} the exact answer. No explanations, no sentences, no extra words.\\[4pt]

\textbf{STRICT OUTPUT RULES:}
\begin{itemize}
    \item ONE word or short phrase only
    \item NO sentences or explanations
    \item NO "This is..." or "The image shows..."
    \item NO punctuation unless part of the answer
\end{itemize}
\end{tcolorbox}
\end{center}

\newpage
\noindent \textbf{Chain-of-Thought Prompt}
\begin{center}
\begin{tcolorbox}[colback=gray!5!white, colframe=black, title={CoT Prompt for Bangla Med-VQA}, width=0.49\textwidth]
\small
You are an expert medical vision-language assistant.\\
\textbf{Task:} You will be given a medical image and a single question about that image (e.g., modality, organ, abnormality, or specific finding). Your job is to think and reason step by step internally using the process below, then provide only the final answer without showing your reasoning.\\[6pt]

\textbf{Step-by-step internal reasoning (do not output this):}
\begin{enumerate}
    \item \textbf{Image Type Identification:}
    \begin{itemize}
        \item Identify the imaging modality (X-ray, CT, MRI, ultrasound, etc.)
        \item Note the anatomical region or body part being examined
    \end{itemize}

    \item \textbf{Visual Analysis:}
    \begin{itemize}
        \item Observe anatomical structures and overall appearance
        \item Check for abnormalities, lesions, devices, or unusual findings
        \item Consider image quality, positioning, and technical factors
    \end{itemize}

    \item \textbf{Clinical Context Assessment:}
    \begin{itemize}
        \item Recall the expected normal appearance for this view
        \item Identify deviations from normal and their significance
    \end{itemize}

    \item \textbf{Question-Specific Reasoning:}
    \begin{itemize}
        \item Link the visual findings directly to the question asked
        \item Consider differential diagnoses only if needed to answer
    \end{itemize}

    \item \textbf{Evidence-Based Conclusion:}
    \begin{itemize}
        \item Decide the most accurate answer supported by the image
        \item Acknowledge uncertainty if evidence is insufficient
    \end{itemize}
\end{enumerate}

\textbf{Output instructions (what to output):}\\
CRITICAL: You must respond with \textbf{ONLY} the exact answer. No explanations, no sentences, no extra words.\\[4pt]

\textbf{STRICT OUTPUT RULES:}
\begin{itemize}
    \item ONE word or short phrase only
    \item NO sentences or explanations
    \item NO "This is..." or "The image shows..."
    \item NO punctuation unless part of the answer
\end{itemize}

\textbf{Remember:} Respond with ONLY the answer, nothing else.
\end{tcolorbox}
\end{center}

\end{document}